\documentclass[sigconf]{acmart}
\usepackage{soul}
\usepackage{multirow}
\usepackage{pifont}
\usepackage{xcolor}
\usepackage{graphicx}
\usepackage{booktabs}
\usepackage{algorithm}
\usepackage{algpseudocode}
\usepackage{hyperref}
\usepackage{subfigure}

\definecolor{lightgreen}{rgb}{0.7137, 0.8431, 0.6588}
\sethlcolor{lightgreen}
\definecolor{darkblue}{rgb}{0.0, 0.0, 134}
\AtBeginDocument{%
  \providecommand\BibTeX{{%
    \normalfont B\kern-0.5em{\scshape i\kern-0.25em b}\kern-0.8em\TeX}}}

\setcopyright{acmlicensed}
\copyrightyear{2024}
\acmYear{2024}
\acmDOI{XXXXXXX.XXXXXXX}

\acmConference[Conference acronym 'MM]{Proceedings of the 32st ACM International Conference on Multimedia}{October 28--November 01, 2024}{Melbourne, Australia}
\acmISBN{978-1-4503-XXXX-X/18/06}

\begin{document}

\title{Game-MUG: Multimodal Oriented Game Situation Understanding and Commentary Generation Dataset}

\author{Zhihao Zhang}
\affiliation{%
  \institution{University of Sydney}
  \country{Australia}
}
\email{zhihao.zhang1@sydney.edu.au}

\author{Feiqi Cao}
\affiliation{%
  \institution{University of Sydney}
  \country{Australia}
}
\email{fcao0492@uni.sydney.edu.au}

\author{Yingbin Mo}
\affiliation{%
  \institution{University of Sydney}
  \country{Australia}
}
\email{yimo6410@uni.sydney.edu.au}

\author{Yiran Zhang}
\affiliation{%
 \institution{University of Sydney}
 \country{Australia}
}
\email{yzha5806@uni.sydney.edu.au}

\author{Josiah Poon}
\affiliation{%
  \institution{University of Sydney}
  \country{Australia}
}
\email{josiah.poon@sydney.edu.au}

\author{Caren Han}
\affiliation{%
  \institution{University of Melbourne}
  \country{Australia}
}
\email{caren.han@unimelb.edu.au}

\renewcommand{\shortauthors}{Zhang, et al.}

\begin{abstract}
The dynamic nature of esports makes the situation relatively complicated for average viewers. Esports broadcasting involves game expert casters, but the caster-dependent game commentary is not enough to fully understand the game situation. It will be richer by including diverse multimodal esports information, including audiences' talks/emotions, game audio, and game match event information. 
This paper introduces GAME-MUG, a new multimodal game situation understanding and audience-engaged commentary generation dataset and its strong baseline. Our dataset is collected from 2020-2022 LOL game live streams from YouTube and Twitch, and includes multimodal esports game information, including text, audio, and time-series event logs, for detecting the game situation. In addition, we also propose a new audience conversation augmented commentary dataset by covering the game situation and audience conversation understanding, and introducing a robust joint multimodal dual learning model as a baseline. We examine the model's game situation/event understanding ability and commentary generation capability to show the effectiveness of the multimodal aspects coverage and the joint integration learning approach.
\end{abstract}



\received{20 February 2007}
\received[revised]{12 March 2009}
\received[accepted]{5 June 2009}

\maketitle

\section{Introduction}
The recent advent of esports has led to a trendy and rapidly growing industry, capturing the attention of a large and continuously expanding global audience. Within a few seconds of a game event, numerous aspects demand attention, such as player action, skills demonstrations, team cooperation, gain and loss, and the key items contributing to the specific game events. This requires the audience to quickly digest complicated information whenever something significant happens in the game. Unlike conventional sports broadcasting like NBA games~\cite{yu2018fsn}, where the fundamental sport's concepts are easily comprehensible, this dynamic nature of esports introduces complexity, making it challenging for the average audience to grasp the game situation fully. Therefore, we need to find a way to assist the audience in understanding the game situation better.
Esports competition organisers address this issue by involving one or two casters to explain the game situation during live streaming. However, this heavily relies on the specific casters, making it difficult for them to provide more diverse information, including audience opinions, feelings, and detailed game match information. In addition, different casters may prioritise various game aspects, leaving many online esports game resources unexplained. Therefore, it is essential to explore methods for automatically generating game-related commentary that comprehensively understands the game situation, incorporating multiple aspects, such as audience discussion, emotions, and domain-specific information though fusing multi-modal features is still quite challenging \cite{cao2022understanding}.

Existing esports game commentary datasets~\cite{LoL-V2T,wang2022esports,zhang-etal-2022-moba} only utilise single-modal information as input to generate textual commentary, disregarding the potential richness of multiple aspects that can provide valuable information about the game. The lack of multimodal resources hinders researchers interested in commentary generation for Multiplayer Online Battle Arena (MOBA) games from determining the best approach to leverage information from various sources to address the game commentary task. Moreover, previous works primarily focus on providing accurate game-related facts~\cite{wang2022esports,zhang-etal-2022-moba} in the generated commentary for the audience, neglecting the importance of infusing human-like qualities and emotions to engage the audience better. Due to the lack of resources, existing game commentary generation models~\cite{LoL-V2T,zhang-etal-2022-moba,wang2022esports} simply employ an encoder-decoder to process raw game information and generate human-like commentary without fully understanding the game situations. 

\begin{table*}[t]
\centering
\label{tab:game-datasets}
\scalebox{1}{%
\begin{tabular}{cccc}
\toprule
\textbf{Dataset} &
  \textbf{\# Matches} &
  \textbf{Modality sources} &
  \textbf{Core Task}\\ \midrule
\textbf{FSN~\cite{yu2018fsn}}            & 50    & video, transcript & Game commentary generation    \\
\textbf{Getting Over It~\cite{li2019gettingover}}            & 8    & video, audio, transcript & Game commentary generation    \\
\textbf{Minecraft~\cite{shah2019automated}}            & 3    & video, transcript & Game commentary generation     \\
\textbf{MOBA LoL~\cite{ringer2019mobalol}}           & -    & video, audio, streamer's image & Streamer emotion prediction, game event type prediction         \\ 
\textbf{Car Racing~\cite{ishigaki-etal-2021-generating}}            & 1,389    & video, game info, transcript & Game commentary generation           \\ 
\textbf{LoL-V2T~\cite{LoL-V2T}}            & 157   & video, transcript & Game commentary generation          \\ 
\textbf{eSports Data-to-Text~\cite{wang2022esports}}            & -   & game info, transcript & Game commentary generation          \\ 
\textbf{Dota2-Commentary~\cite{zhang-etal-2022-moba}}        & 234    & game info, transcript & Game commentary generation          \\ 
\textbf{CS-lol~\cite{xu2023cslol}}             & 20    & transcript, chat & Viewer comment retrieval \\ 
\midrule
\textbf{Game-MUG (ours)}            & 216    & audio, chat, game info, transcript & Game commentary generation, game event type prediction     \\
\bottomrule

\end{tabular}%
}
\caption{The summary of existing game datasets and the comparison of our proposed dataset}
\label{tab:game_datasets}
\end{table*}

We introduce GAME-MUG, a multimodal game situation understanding and commentary generation dataset, and its strong baseline. Our dataset incorporates publicly available League of Legends (LOL) resources with professional caster comments from popular live streaming platforms, YouTube and Twitch, with multimodal information, including game event logs, caster's speech audio, and game-related natural language discussions encompassing both human casters' speech and audience chats and emotions. Inspired by the joint integration of natural language understanding and generation tasks, we propose a strong baseline model that employs joint integration framework to comprehend game situations from multimodal information and generate game commentary based on this understanding of game situations and emotions. To conduct the game commentary generation, we summarise the game situation and audience conversation via multi-modality sources.

The contribution of this paper can be summarised as follows:
\begin{itemize}
    \item We introduce a multimodal game understanding and commentary generation dataset to provide a full understanding of the game situations with caster comments and diverse information, including audience conversation, caster's speech audio, and game event logs.
    \item We propose a joint integration framework to generate more human-like commentary with the help of game situation understanding
    \item We conduct extensive experiments to show the effectiveness of multimodality in game understanding and commentary generation.
\end{itemize}

\section{Related Work}

\subsection{Game-related Datasets}
Most datasets in the game domain are proposed for commentary generation across different games, such as live-streamed MOBA games~\cite{LoL-V2T,wang2022esports,zhang-etal-2022-moba} as well as pre-recorded esports games~\cite{ishigaki-etal-2021-generating,li2019gettingover,shah2019automated} or traditional sports~\cite{yu2018fsn}, while several datasets also focus on classification tasks related to scene understanding as shown in Table~\ref{tab:game_datasets}. CS-lol~\cite{xu2023cslol} proposed a task of viewer comment retrieval, while MOBA-LoL~\cite{ringer2019mobalol} proposed two classification tasks on their dataset. On top of predicting game event types, they also provide multi-view to understand the game context, by indicating the streamer's emotional state.
Among all the datasets proposed for game commentary generation, most datasets allow only a single modality as the input, video only, or game information only. Some datasets allow multimodal input, but it was not for MOBA games. So far, no previous work utilises audience emotion when they build datasets to generate more human-like commentary for MOBA games. Our dataset provides both audience emotion and rich multimodal input, including audio, audience chat, and game information.

\subsection{Visual-Linguistic Generation}
Most works in visual-linguistic generation tasks like video captioning or commentary generation for games used encoder-decoder structure~\cite{yu2018fsn,li2019gettingover,shah2019automated,ishigaki-etal-2021-generating,LoL-V2T,zhang-etal-2022-moba,wang2022esports}, and some~\cite{LoL-V2T,zhang-etal-2022-moba,wang2022esports,cao2023scenegate} experimented with several types of structures like unified encoder-decoder, pretraining method, rule-based models, and hybrid models. Some works~\cite{li2019gettingover,wang2022esports,zhang-etal-2022-moba,ishigaki-etal-2021-generating,yu2018fsn} applied recurrent seq2seq models like LSTM/GRU structures for both encoding the input and decoding for commentary, some~\cite{LoL-V2T,wang2022esports,zhang-etal-2022-moba} used transformer-based models for generating commentary. However, no model used dense interaction/fusion among different input modalities. Previous models either lack multimodal input or concatenate different modality features as one feature vector or via simple tensor operation. The semantic gap between different modalities is ignored. In addition, no work tried dual learning of understanding game scenes and generating commentary due to limited information provided by datasets. Our method uses the audience's chats and opinions to understand the game context to facilitate the automatic generation of commentary.

\section{Definitions}
In esports broadcasting, elucidating the intricacies of gameplay dynamics is essential for audience engagement. Traditional methods rely on casters to provide real-time commentary during live streams. However, this approach has limitations, leading to the emergence of alternative solutions such as our proposed commentary system. In this subsection, we define \textit{caster's speech} and contrast it with the features and benefits of \textit{our proposed game situation-based commentary}. The rest of the paper would consistently use the following terms to help readers' understanding.

\noindent\textbf{Caster's Speech}
Esports competitions commonly employ one or two casters to articulate the ongoing game situation. These individuals play a pivotal role in providing context and analysis to viewers. However, the caster-dependent commentary is rather subjective and heavily contingent on the expertise and style of the specific casters involved.

\noindent\textbf{Our proposed Commentary}
Our proposed commentary system addresses the shortcomings of caster-dependent approaches by offering a more comprehensive and diverse narrative. Unlike traditional commentary, our system incorporates multimodal elements, including online game audience sentiments, real-time game audio, and detailed game match event information. By integrating audience conversations and prioritising inclusivity, our proposed commentary comprehensively understands the game match event and audience sentiment.

\section{Game-MUG}
We introduce a new game commentary dataset using multimodal game situational information, called Game-MUG. It features three modalities: game match event logs, audio features derived from signal data and textual discussions, such as caster's speech transcript and audience chat. It comprises 70k clips with transcripts and 3.7M audience chats collected from 45 LOL competition live streams. Each live stream has an average of 4.8 individual matches, leading to 216 game matches and 15k game events. Game matches are sourced from 3 distinct leagues between 2020 and 2022, including \href{https://lolesports.com/standings/lpl}{Tencent League of Legends Pro League}, \href{https://lolesports.com/standings/lck}{League of Legends Champions Korea} and \href{https://lolesports.com/standings/worlds}{World Championships}. These top-tier league matches in various regions attract many views (from 507K to 7.2M), which derives abundant audience chats in multiple languages. We collect caster's speech and audience live chats from two different livestream platforms: \href{https://www.twitch.tv/}{Twitch}, which contributes 150 matches, and \href{https://www.youtube.com/@live}{YouTube}, which contributes 66 matches. In addition to this, we crawl game events from the League of Legends Competitive Statistics Website\footnote{\url{https://gol.gg/esports/home/}}. 

\subsection{Data Collection}
\textbf{Gaming Human Commentary Transcription.} We collect human caster's speech by transcribing the raw live stream files\footnote{YouTube and Twitch disable their Automatic Speech Recognition tools on game live streams}. Due to the substantial size of live-stream videos, we use \href{https://github.com/yt-dlp/yt-dlp}{YT-DLP} and \href{https://github.com/ihabunek/twitch-dl}{Twitch-DL} only to download their high-definition (44.1kHz) audio and utilise a speech recognition model named Whisper~\cite{whisper} for speech-to-text conversion. Whisper is a large supervised model that implies the encoder-decoder architecture from Transformer~\cite{vaswani2017attention}. We use Whisper \href{https://github.com/openai/whisper}{medium English model} and set the compression ratio to 1.7 without previous text conditions for speech-to-text recognition, which slightly trades off the transcript accuracy but maximises its robustness. Each transcribed text is paired with its start and end timestamps in seconds.

\textbf{Audience Live Chats Collection.} Audience live chats are scrapped from the live stream platforms. We employ a multiplatform software named \href{https://github.com/xenova/chat-downloader}{Chat Downloader} to scrap the chat content from YouTube and Twitch. Because of the multilingual nature of live chats, we use \href{https://github.com/pemistahl/lingua}{Lingua} to identify different languages and apply a special label called “emo” for chat instances that only include emotes or emojis. Live stream platforms automatically filter out hateful and toxic contents and we further filter out the live chats without any content and associate remainders with their respective timestamps in seconds.
To ensure the anonymity and privacy of individuals involved in the live chats, we implemented a de-identification protocol. The primary objective of this protocol is to mask any information that could potentially reveal the identity of a chat participant. We directly remove all original usernames associated with the chats, ensuring it is infeasible to reverse engineer the original usernames. All de-identified chats are stored in plain text format, without any identifying information. The original raw data are permanently deleted after the de-identification process. By taking these steps, we ensure that our data collection and analysis processes align with ethical guidelines and data protection regulations.

\textbf{Game Events Collection.} Game events are collected from the League of Legends Competitive Statistics Website by a scrapper; it first finds the game-related HTML tags and extracts the contents from the selected tags. It is worth noticing that sometimes the contents of the tags can be empty, which means a minion or a non-epic monster triggers this event. Our scrapper automatically populates missing contents in the tags and links them to game timestamps, constructing complete game event instances. We categorise collected game events into the following 6 different classes in our dataset: \textbf{1) Kill:} A game character is defeated; \textbf{2) Non-Epic Monster:} A jungle monster is eliminated; \textbf{3) Tower:} A turret/inhibitor is destroyed; \textbf{4) Dragon:} A dragon is eliminated; \textbf{5) Plate:} A turret’s defensive barrier is shattered; \textbf{6) Nexus:} An nexus is destroyed, leading to the end of the game.

\textbf{Audio Feature Extraction.} It is known that human speech tone fluctuates based on emotions~\cite{kienast00_speechemotion} and audio modality demonstrates a notable advantage over video in capturing emotional fluctuations~\cite{10.3389/fcomp.2021.767767}. Therefore, we extract audio features from the caster's speech audio to enrich emotional representation within diverse domain data. The Geneva Minimalistic Acoustic Parameter Set (GeMAPS)~\cite{7160715} is commonly used for voice research and it encompasses 18 Low-Level Descriptors, which covers features related to frequency, amplitude and spectral parameters. We utilise \href{https://github.com/audeering/audiofile/}{audiofile} to convert raw audio files into audio waveforms, and then extract audio features with a sampling rate of 50Hz using openSMILE~\cite{10.1145/1873951.1874246}, a tool commonly used for vocal emotion recognition~\cite{opensmile-ver}.

\begin{table}[]
\centering
\resizebox{0.35\textwidth}{!}{%
\begin{tabular}{@{}cccc@{}}
\toprule
\textbf{Categories} & \multicolumn{1}{c}{\textbf{GPT-3.5}} & \multicolumn{1}{c}{\textbf{GPT-4}} & \multicolumn{1}{c}{\textbf{Tie}}  \\ \midrule
\textbf{Kill}    & 25.78\% & 51.56\% & 22.66\% \\
\textbf{Tower}   & 14.20\% & 59.66\% & 26.14\% \\
\textbf{Dragon}  & 17.71\% & 66.67\% & 15.63\% \\ \midrule
\textbf{Overall} & 18.75\% & 58.75\% & 22.50\% \\ \bottomrule
\end{tabular}%
}
\caption{Pairwise comparison between GPT-3.5 and GPT-4 commentaries, the overall agreement coefficient~\cite{Krippendorff} is 0.64 from nine human annotators. In most cases, annotators choose GPT-4 summaries over GPT-3.5 or think they are similar.}
\label{tab:label-human-eval}
\end{table}

\subsection{Data Annotation}

\textbf{Game Situation Commentary Annotation.} Inspired by the success of Standford Alpaca~\cite{alpaca}, we make use of GPT-3.5~\cite{gpt-3-5} and GPT-4~\cite{openai2023gpt4} to condense all 70,711 human caster's speech transcripts into concise commentaries with emotional clues from audience chats as detailed in Algorithm~\ref{alg:prompts}.We set the background information as watching a live game streaming via a system prompt. Whenever a game event occurs, we forward the caster's speech transcript and live chat content to the GPT-4 API through the commentary prompts. We design several prompt parameters to guide the GPT-4 generation: \textcolor{darkblue}{\textless game streaming platform\textgreater} indicates different live stream platforms, \textcolor{darkblue}{\textless number of commentary words\textgreater} control the number of generated words, and \textcolor{darkblue}{\textless game-related topics\textgreater} adjusts the generated commentary to focus on different aspects, such as on player, character, event or overall situation. 
To ensure the annotation quality, we conduct a pairwise human evaluation between the commentaries generated from GPT-3.5 and GPT-4. As shown in Table~\ref{tab:label-human-eval}, GPT-4 excels GPT-3.5 in all three categories, indicating GPT-4's commentaries are better aligned with human understanding. Therefore, we choose GPT-4's commentaries as ground truth annotations in our dataset.

\begin{algorithm*}
\caption{Game Situation Commentary Annotation}
\label{alg:prompts}
\begin{algorithmic}
\Require \textcolor{darkblue}{\textless game streaming platform\textgreater}, \textcolor{darkblue}{\textless number of commentary words\textgreater}, \textcolor{darkblue}{\textless game-related topics\textgreater}
\Ensure Input \hl{caster's speech transcripts and audience chat}

\Statex
\Procedure{Background Information}{}
    \State \parbox[t]{\dimexpr\linewidth-1.5em}{\textbf{{System Prompt:}} You are watching the League of Legends Competition live stream from \textcolor{darkblue}{\textless game streaming platform\textgreater} with other audiences.}
\EndProcedure

\Statex
\Procedure{game situation commentary annotation}{}
    \State \parbox[t]{\dimexpr\linewidth-1.5em}{\textbf{Summary Prompt:} Based on the \textcolor{darkblue}{\textless system prompt\textgreater}, generate a one-sentence commentary between \textcolor{darkblue}{\textless number of commentary words\textgreater} from this \hl{caster's speech transcript} highlighting \textcolor{darkblue}{\textless game-related topics\textgreater}, while incorporating the audience's emotions from this \textcolor{darkblue}{\textless game streaming platform\textgreater} \hl{audience chat}.}
\EndProcedure

\end{algorithmic}
\end{algorithm*}

\begin{table}[]
\centering
\resizebox{0.45\textwidth}{!}{%
\begin{tabular}{cccc}
\toprule
\textbf{Event} &
  \textbf{\begin{tabular}[c]{@{}l@{}}\# of events\end{tabular}} &
  \textbf{\begin{tabular}[c]{@{}l@{}}Avg per match \end{tabular}} &
  \textbf{\begin{tabular}[c]{@{}l@{}}Percentage \end{tabular}} \\ \midrule
\textbf{Kill}             & 5,548   & 25.69 & 36.45\% \\ 
\textbf{Tower}            & 2,889   & 13.38 & 18.98\% \\ 
\textbf{Dragon}           & 1,646   & 7.62  & 10.81\% \\  
\textbf{Other}            & 5,138   & 23.79 & 33.76\%          \\ \midrule
\textbf{Total}            & 15,221  & 70.47 & 100\%            \\ \bottomrule
\end{tabular}%
}
\caption{Distributions of the more important game events in our collected dataset, where the less important ones, \textbf{Non-Epic Monster}, \textbf{Plate} and \textbf{Nexus}, are categorised into \textbf{Other} as an initial step for analysis.}
\label{tab:table-event}
\end{table}

\subsection{Data Processing}
Considering each live stream can be treated as a chronological sequence comprised of game events, caster's speech and live chats, we match them via their timestamps.  As game events’ timestamps are reset after each match, we manually adjust them to align with live stream seconds prior to the matching process. Additionally, background music before the commencement of each live stream is also removed manually, since there is no game-related factual information to help with game situation understanding. 

\section{Data Analysis}\label{sec:analysis}
Our dataset includes 70,711 transcript sentences with an average duration of 12.2 secs and 3,657,611 chat instances. 15,221 game events are collected from 216 game matches. Not all events are equally important for the human caster and audience; \textbf{Kill}, \textbf{Tower}, and \textbf{Dragon} events usually attract more interest than other events. Therefore, we categorise all other events into \textbf{Other} as an initial input processing step for our following analysis in Section~\ref{sec:analysis} and experiments in Section~\ref{sec:results}. We present the statistics of each event category in Table~\ref{tab:table-event}.

\subsection{Game Keyword Analysis}

\begin{figure}[h]
  \centering
  \includegraphics[width=0.48\textwidth]{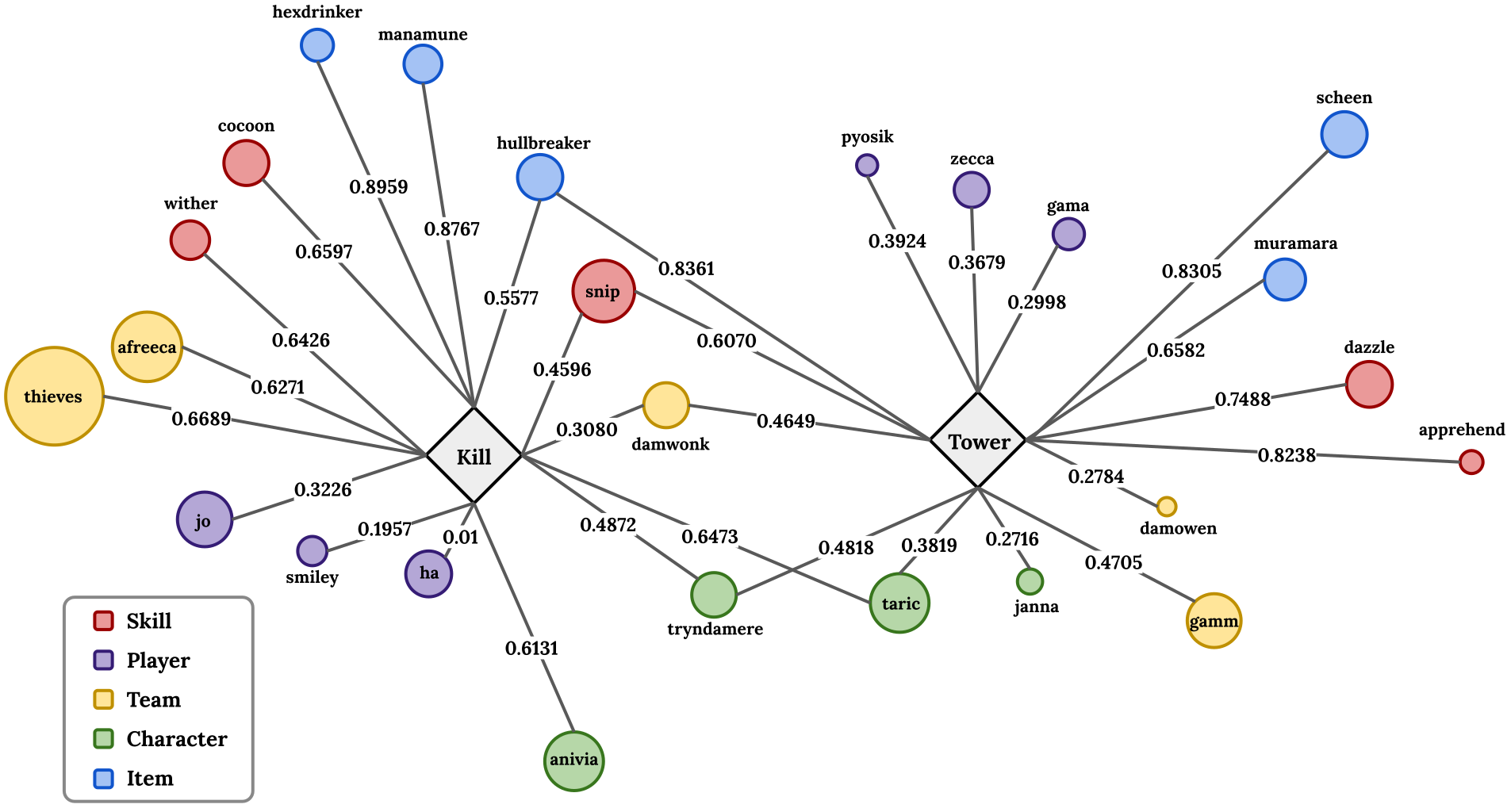}
  \caption{The visualisation for keyword analysis with top 15 words from \textbf{Kill} and \textbf{Tower} Event, where the time window is 30s. Entities related to each event, Kill and Tower, are remarkably different, such as skill `cocoon` for \textbf{Kill} and `apprehend` for \textbf{Tower}.}
  \label{fig:keyword-graph}
\end{figure}

Different from other domains, game-related data contains numerous keywords that rarely appear in everyday conversations. We manually extract 2,003 unique keywords from the caster's speech transcript in our dataset and clean the typos and misspells while retaining essential abbreviations, such as character’s skills denoted by Q, W, E, and R. As shown in Figure~\ref{fig:keyword-graph}, extracted keywords can be categorised into 5 different classes, including skill, player, team, character and item. 
To better address the importance of each keyword, we compute their Term Frequency - Inverse Document Frequency (TF-IDF) based on the game events with different time windows, specifically 15 seconds and 30 seconds. The transcripts encompassed within these windows are treated as a singular document to compute TF-IDF values. This allows us to identify key terms closely associated with game events. Depending on the precise timing of the event, such a window might encapsulate one or several caster's speech transcripts. This calculation is performed using the Scikit-learn library~\cite{scikit-learn} with normalisation. Figure~\ref{fig:keyword-graph} shows a sample visualisation of the keywords' characteristics when the window of time equals 30 seconds. We select the top 15 keywords for \textbf{Kill} and \textbf{Tower} events and differentiate their types by distinct colours. The size of each keyword’s node depends on the normalised occurrence of the keyword, whereas the distance between the event and keyword nodes is determined by the normalised TF-IDF values. From Figure~\ref{fig:keyword-graph} we can see that \textbf{Kill} and \textbf{Tower} are more related to items to attack, skills that either increase the damage for attacking enemies or limit the ability of enemies moving to avoid damage or fighting back. This reflects the typical player's actions in games, which often involve attacking opponents, indicating that the text in our dataset effectively describes the game scene and offers a robust understanding of the situation. Moreover, we can see that team, players, and character names are frequently mentioned or discussed by commentators when these cases happened; though the names might depend on specific games, it demonstrates the multiple aspects people could focus on about the game situation.

\begin{figure}[t]
  \centering
  \includegraphics[width=0.48\textwidth]{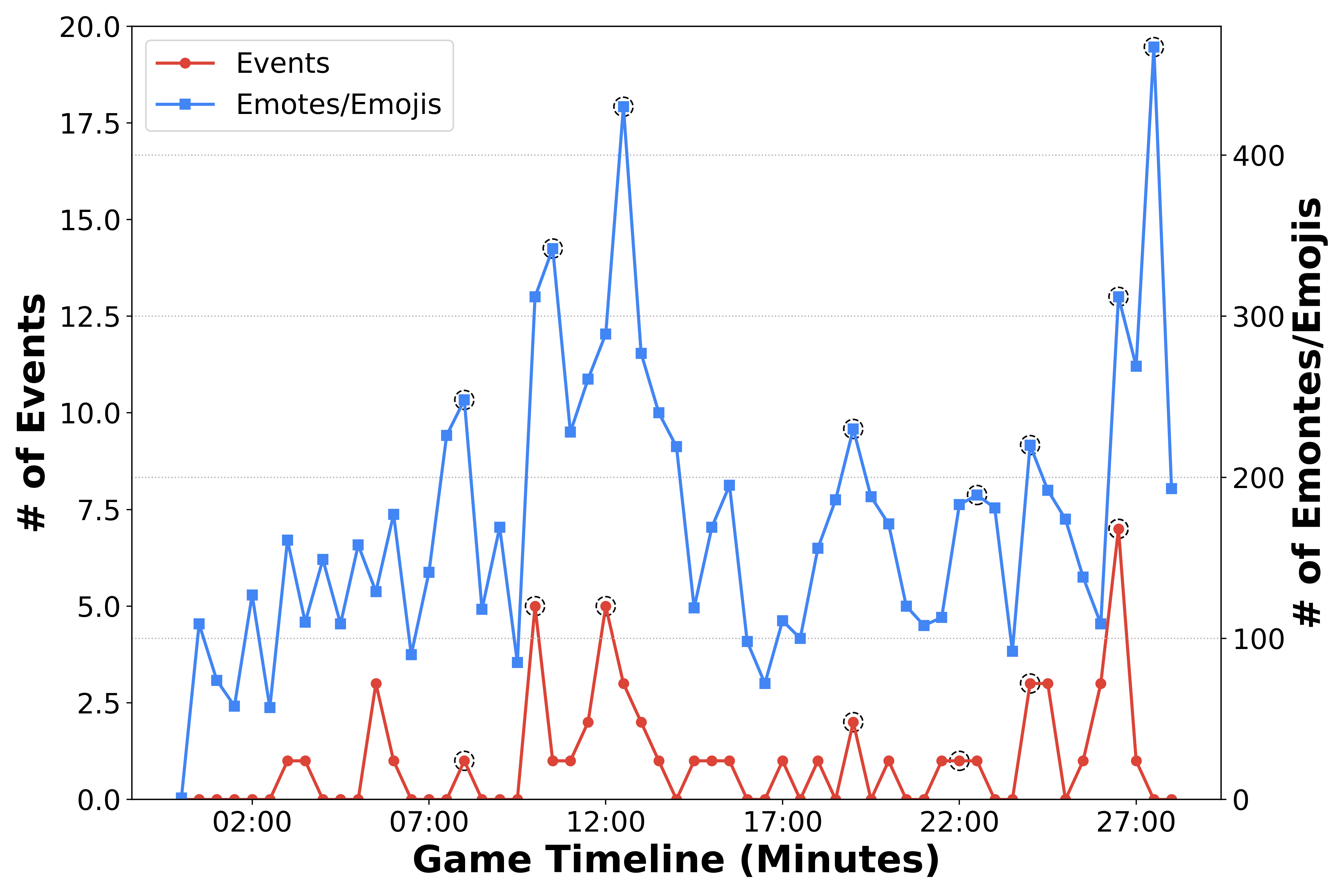}
  \caption{The concurrent plot for audience chat analysis with the numbers of emotes, emojis, and game events along the same timeline. A positive correlation can be observed between the number of audience chat emojis and the number of game events happening within the same time window.}
  \label{fig:emo}
\end{figure}

\subsection{Audience Chat Analysis}

The audience tends to send many emotes and emojis in chat to express their sentiments. We retrieve emotes and emojis based on their distinct formats found in publicly available sources\footnote{\url{https://www.frankerfacez.com/emoticons/}}\footnote{\url{https://github.com/carpedm20/emoji/}} and then count the number of emotes and emojis per 30-second window in each match. The counts of emotes, emojis, and game events are plotted concurrently on the same timeline, shown in Figure~\ref{fig:emo}. It is not hard to discover that the number of emotes correlates with the game situation, since audiences tend to send more emotional expressions in chats to share their feelings when a dramatic turning point or a series of events happens.

\section{Proposed Baseline}
Based on Game-MUG, we proposed a joint integration framework that generates commentaries based on understanding the game situation through multimodal data. For game situation understanding, we implemented and fine-tuned a multimodal transformer encoder that encodes text and audio data. For game commentary generation, we employ a pre-trained decoder and encoded game information. The quality of generated commentaries is evaluated by both automatic metrics and humans. We partition our dataset into 206 matches for training and 10 matches for testing. 

\begin{figure*}[h]
  \centering
  \includegraphics[width=1\textwidth]{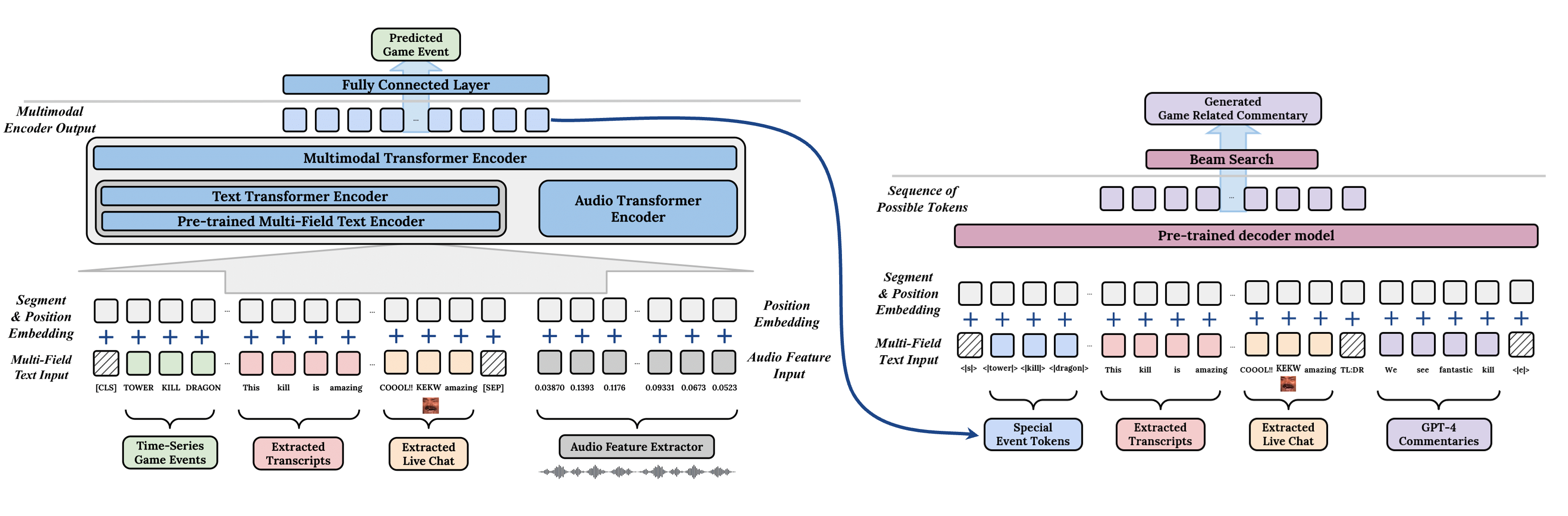}
  \caption{Joint integration framework of Game Situation Understanding and Game Commentary Generation}
  \label{fig:overall-structure}
\end{figure*}

\subsection{Input Processing}
Given an $i$-th event $E_i$ happening at $t_{ei}$ of a game, we try to predict its event type via the multimodal information provided in our dataset and the game situation understanding module, and generate a commentary via the game commentary generation module. Taking $m$ most recent game events which happened before $E_i$ as a historical reference, we extract the time-series event sequence as $\mathbb{E}=\{E_{i-m}, …, E_{i-2}, E_{i-1}\}$. Assuming that the input window size for transcript and chat is $w$, we extract a time-series sequence consisting of $x$ transcript clips $\mathbb{T}=\{T_{s-x}, …, T_{s-1}, T_{s}\}$, where $T_{s}$ refers to the $s$-th transcript clip in the current game. These clips fully cover the time period from $(t_{ei}-w)$ to $t_{ei}$, meaning that the timestamp $(t_{ei}-w)$ falls within the time frame covered by $T_{s-x}$, and $t_{ei}$ falls within the time frame covered by $T_{s}$. The time-series sequence of chats $\mathbb{C}$ is extracted based on their specific timestamps between $(t_{ei}-w)$ and $t_{ei}$. For the audio component, given the window size $w_a$, the audio feature sequence is extracted as $\mathbb{A}$ within the time period between $(t_{ei}-w_a)$ and $t_{ei}$. This results in a vector consisting of $w_a*50$ values that serve as the input for the audio transformer, given that the audio features are sampled at a rate of 50Hz.

\subsection{Game Situation Understanding}
The model architecture is shown on the left of Figure~\ref{fig:overall-structure}. On the text side, the input is a combination of multi-field sequential time-series data from previous event $\mathbb{E}$, caster's speech transcript $\mathbb{T}$ and audience chat $\mathbb{C}$, with graphical emotional expressions in chats being converted into their text representation. Since chats tend to contain many repetitions in phrases and emotions, we truncate the input sequence up to 256 tokens. Following the approaches in BERT~\cite{devlin-etal-2019-bert}, we insert a [CLS] token at the beginning and a [SEP] token at the end of the input sequence, creating the input embeddings by summing the token, segment, and position embeddings. These input embeddings are initially passed into a pre-trained multi-field text encoder. The [CLS] token output from this pre-trained multi-field text encoder is then forwarded to the text transformer encoder to project the text representation into a common space. On the audio side, the combination of audio feature $\mathbb{A}$ and position embedding are fed into an audio transformer, which maps the audio into the same common space as the text. The text and audio representations are then concatenated to form a single vector, which serves as the input for the multimodal transformer encoder followed by a fully connected layer to predict the subsequent game event. We take advantage of existing pre-trained models in our multi-field text encoder including BERT~\cite{devlin-etal-2019-bert}, RoBERTa~\cite{liu2019roberta}, DeBERTa~\cite{he2021deberta}, and XLNet~\cite{NEURIPS2019_xlnet}. More details can be found in Section~\ref{sec:eval_setup}.

\subsection{Game Commentary Generation}
After fine-tuning the game situation understanding model, we obtain the event representations from it before the fully connected layer and incorporate these representations along with transcripts and chats into the pre-trained generative model for commentary generation. We calculate the mean of each event representation by inference the trained game situation understanding model with all the matches in our dataset to get the special event embeddings. These embeddings are then added to the decoder models’ vocabulary as <|kill|>, <|tower|> and <|dragon|> to enhance efficiency during commentary generation. Similar to the encoder model, we truncate the chat sequence up to 256 tokens for emotion extraction before combination them with special event tokens and transcripts. As shown in the right of Figure~\ref{fig:overall-structure}, a special [TL;DR] token and GPT-4 commentaries are concatenated to the sequence as a reference during fine-tuning. We utilise two different pre-trained decoders, including GPT-2~\cite{radford2019language} and Pythia~\cite{biderman2023pythia}. More details can be found in Section~\ref{sec:eval_setup}.

\begin{table*}[]
\centering

\resizebox{\textwidth}{!}{%
\begin{tabular}{ccc|cccc|cccc|cccc|cccc}
\hline
\multirow{2}{*}{\textbf{Chat}} & \multirow{2}{*}{\textbf{Audio}} & \multirow{2}{*}{\textbf{\begin{tabular}[c]{@{}l@{}}Game\\ Events\end{tabular}}} & \multicolumn{4}{c|}{\textbf{BERT}} & \multicolumn{4}{c|}{\textbf{DeBERTaV3}} & \multicolumn{4}{c|}{\textbf{RoBERTa}} & \multicolumn{4}{c}{\textbf{XLNet}} \\ \cline{4-19} 
 &  &  & \textbf{Kill} & \textbf{Tower} & \textbf{Dragon} & \textbf{All} & \textbf{Kill} & \textbf{Tower} & \textbf{Dragon} & \textbf{All} & \textbf{Kill} & \textbf{Tower} & \textbf{Dragon} & \textbf{All} & \textbf{Kill} & \textbf{Tower} & \textbf{Dragon} & \textbf{All} \\ \hline
\multicolumn{1}{c}{\ding{56}} & \multicolumn{1}{c}{\ding{56}} & \multicolumn{1}{c|}{\ding{56}}  & 77.98 & 47.75 & 8.45 & 61.97 & 79.46 & 62.16 & 1.41 & 65.06 & 79.17 & 62.16 & 8.45 & 65.83 & 93.75 & 10.81 & 4.23 & 63.71 \\
\multicolumn{1}{c}{\ding{52}} & \multicolumn{1}{c}{\ding{56}} & \multicolumn{1}{c|}{\ding{56}} & 86.01  & 20.72  & 9.86  & 61.58  & 81.55  & 62.16 & 0.00  & 66.22  & 79.46  & 59.46  & 7.04  & 65.25  & 96.43  & 0.90  & 5.63  & 63.51  \\
\multicolumn{1}{c}{\ding{56}} & \multicolumn{1}{c}{\ding{52}} & \multicolumn{1}{c|}{\ding{56}} & 83.63  & 37.84  & 14.08  & 64.29  & 77.08  & 36.94  & 49.30  & 64.67  & 78.57  & 62.16 & 25.35  & 67.76  & 72.02  & 55.86  & 22.54  & 61.78  \\
\multicolumn{1}{c}{\ding{56}} & \multicolumn{1}{c}{\ding{56}} & \multicolumn{1}{c|}{\ding{52}} & 80.55  & 51.35  & 17.19  & 64.96  & 72.35  & 61.26  & 17.19  & 62.18  & 78.50  & 58.56  & 35.94  & 67.95  & 95.22  & 15.32 & 0.00  & 63.25  \\
\multicolumn{1}{c}{\ding{52}} & \multicolumn{1}{c}{\ding{52}} & \multicolumn{1}{c|}{\ding{56}} & 75.00  & 48.65  & 11.27  & 60.62  & 82.44  & 43.24  & 43.66  & 68.73  & 77.38  & 63.06  & 15.49  & 65.83 & 67.86  & 53.15  & 14.08  & 57.34  \\
\multicolumn{1}{c}{\ding{52}} & \multicolumn{1}{c}{\ding{56}} & \multicolumn{1}{c|}{\ding{52}} & 83.22  & 51.35  & 18.03  & 66.81  & 81.82  & 58.56  & 40.98  & 70.74  & 80.07  & 55.86  & 36.07  & 68.34  & 80.07  & 62.16  & 21.31  & 67.90  \\
\multicolumn{1}{c}{\ding{56}} & \multicolumn{1}{c}{\ding{52}} & \multicolumn{1}{c|}{\ding{52}}  & 84.97  & 32.43  & 42.62  & 66.59  & 79.72  & 49.55  & 34.43  & 66.38  & 84.62  & 51.35  & 26.23  & 68.78  & 76.22  & 60.36  & 21.31  & 65.07  \\
\multicolumn{1}{c}{\ding{52}} & \multicolumn{1}{c}{\ding{52}} & \multicolumn{1}{c|}{\ding{52}} & 83.57  & 43.24  & 18.03  & 65.07  & 86.71  & 31.53  & 59.02  & 69.65  & 80.42  & 52.25  & 31.15  & 67.03  & 83.92  & 53.15  & 18.03  & 67.69  \\ \hline
\end{tabular}%
}
\caption{The effect of Chat, Audio and previous Game Events on 2 different Game Situation Understanding Models.}
\label{tab:understanding-ablation}
\end{table*}

\section{Experiments and Results}

\subsection{Experiment Setup}\label{sec:eval_setup}
\textbf{Game Situation Understanding} We test four pre-trained encoder models with their large settings as the baseline multi-field text encoders: BERT\textsc{large}, RoBERTa\textsc{large}, DeBERTaV3\textsc{large}, and XLNet\textsc{large}. The text and audio transformer encoder and the multimodal transformer encoder are all 8-head and 6-layer encoder structures and 1024 embedding dimension. The entire model is trained using AdamW~\cite{adamw} with 2 epochs for each instance, with a dropout value of 0.1~\cite{dropout}, a learning rate of 1e-6, and a learning rate decay rate of 0.95 for every 2 epochs. 
\textbf{Game Commentary Generation} We adopt two pre-trained decoder models as the baseline commentary generation models: 762M GPT2 with 1280 dimension size and 410M Pythia with 1024 embedding size. We apply Principal Component Analysis~\cite{pca} to the game event embeddings when their dimensions are larger than the embeddings of pre-trained models for fine-tuning consistency. All models are trained using AdamW for 3 epochs, with a learning rate of 1e-5, and a warmup step of 5. 
Our implementations are based on PyTorch~\cite{pytorch} and HuggingFace Transformers~\cite{transformers}, with the help of Scikit-learn~\cite{sklearn_api}. All experiments are run on a test bench with 24GB NVIDIA RTX 3090 GPU.

\subsection{Evaluation Metrics}
We evaluate the game situation understanding model with a multi-class accuracy metric, directly comparing the predicted game event with the ground truth for each event class. Generated commentaries are evaluated with ROUGE~\cite{rouge} and BERTScore~\cite{bertscore}, common automatic evaluation metrics. To have the best correlation with humans, we choose a RoBERTa\textsc{large} version of BERTScore, which deploys a RoBERTa model to compare the similarity between the model generations and references. All results are reported for a single run of the experiments.

\begin{table}[]
\centering
\resizebox{\columnwidth}{!}{%
\begin{tabular}{@{}c|cc|cc@{}}
\toprule
\multirow{2}{*}{\textbf{\begin{tabular}[c]{@{}l@{}}Special Event Token\end{tabular}}} & \multicolumn{2}{c|}{\textbf{GPT2}} & \multicolumn{2}{c}{\textbf{Pythia}} \\ \cmidrule(l){2-5} 
      & BertScore & ROUGE-L & BertScore & ROUGE-L \\ \midrule
\multicolumn{1}{c|}{\ding{56}} & 76.15    & 18.52  & 74.45    & 13.24  \\
\multicolumn{1}{c|}{\ding{52}} & 76.38    & 17.10  & 75.37    & 15.98  \\ \bottomrule
\end{tabular}%
}
\caption{The effect of special event tokens on 2 different Game Commentary Generation Models.}
\label{tab:generation-ablation}
\end{table}

\subsection{Results}\label{sec:results}

\noindent\textbf{Overall Performance}
As illustrated in Table~\ref{tab:understanding-ablation}, when all input features are utilised, DeBERTaV3 notably outperforms the others in overall accuracy as well as \textbf{Kill} and \textbf{Dragon} categories by trading off the performance on \textbf{Tower}. Trailing behind DeBERTaV3, the overall performance of RoBERTa and XLNet is similar, with a margin difference of less than 1\%. It is worth noting that RoBERTa excels in the \textbf{Dragon} category, while XLNet excels in the \textbf{Kill} and \textbf{Tower} categories. Although BERT achieves an overall accuracy of 65.07\%, it ranks last among the four encoder variants. This is likely attributable to the other models' more robust optimisation built upon BERT's architecture. In addition, all models produce better prediction accuracy for \textbf{Kill} than for \textbf{Tower} and \textbf{Dragon}. This trend is primarily due to the imbalanced event data since the average number of \textbf{Kill} instances per match is 25.69, which is double the average number of \textbf{Tower} instances (11.62) and triple the average number of \textbf{Dragon} instances (7.62). Regarding the game commentary generation results presented in Table~\ref{tab:generation-ablation}, we note that GPT2 consistently outperforms Pythia across both evaluation metrics, irrespective of special event tokens.

\noindent\textbf{Ablation Studies}
To further analyse the effectiveness of our data, we conduct ablation studies to compare 3 different input combinations with transcripts for the game situation understanding model: \textbf{1) Audio:} with and without audio features as part of the sequence input; \textbf{2) Chat:} with and without chat as part of the sequence input; \textbf{3) Game Events:} with and without game events as part of the sequence input. The results are presented in Table~\ref{tab:understanding-ablation}. We observed that supplementing the model with additional input data improves its capability for understanding game situations. This results in a noticeable performance increase across all three models, particularly for the rare \textbf{Dragon} event, albeit with a slight trade-off in performance for other events. Specifically, individually incorporating audio or previous game events into the transcript yields a greater improvement than adding chat data alone. Furthermore, combining two types of additional inputs surpasses the performance achieved with just a single extra input.
We also conduct experiments both in the presence and absence of the \textbf{Special Event Token}, defined as the intermediate embedding before the fully connected layer within the game situation understanding model, as illustrated in Figure~\ref{fig:overall-structure}. Other inputs, such as caster's speech transcripts, chats, and GPT-4 commentaries, are essential for fine-tuning since omitting any of these causes a significant drop in generation performance. The results of these experiments are shown in Table~\ref{tab:generation-ablation}. We observed the addition of a special event token can guide model generation, leading to improvements in BertScore for both GPT2 and Pythia.

\begin{table}[]
\centering
\resizebox{\columnwidth}{!}{%
\begin{tabular}{@{}c|c@{\hspace{8pt}}c@{\hspace{5pt}}c@{\hspace{5pt}}c@{\hspace{5pt}}|c@{\hspace{8pt}}c@{\hspace{5pt}}c@{\hspace{5pt}}c@{\hspace{5pt}}@{}}
\toprule
\multirow{2}{*}{\textbf{Audio}} & \multicolumn{4}{c|}{\textbf{BERT}}                              & \multicolumn{4}{c}{\textbf{DeBERTaV3}}                          \\ \cmidrule(l){2-9} 
                       & \textbf{Kill} & \textbf{Tower} & \textbf{Dragon} & \textbf{All} & \textbf{Kill} & \textbf{Tower} & \textbf{Dragon} & \textbf{All} \\ \midrule
5s  & 80.55 & 42.34 & 25.00 & 63.89          & 83.62 & 35.14 & 57.81 & 68.59          \\
10s & 83.62 & 37.84 & 23.44 & 64.53          & 83.28 & 35.14 & 59.38 & 68.59          \\
15s & 84.30 & 40.54 & 18.75 & \textbf{64.96} & 86.35 & 28.83 & 57.81 & \textbf{68.80} \\ \bottomrule
\end{tabular}%
}

\caption{Hyperparameter testing on the Game Situation Understanding Models for different audio time windows (rounded to the nearest integer in order to obtain enough data to match the audio transformer embedding size which should be a multiple of 8), where input transcript and chat time windows are 30s, and the number of previous game events is 5. A larger audio time window may lead to higher performance with a small margin.}
\label{tab:audio-hp-testing}
\end{table}
\noindent\textbf{Hyperparameter Testing}
The audio hyperparameter testing for the three different variations of the Game Situation Understanding Model is in Table~\ref{tab:audio-hp-testing}, where input transcript and chat time windows are set to 30 seconds, and the number of previous game events are set to 5. We observe that the performance of each model is barely influenced by the input length of the audio features, as the difference is within a 1\% margin. 
We also explore the effectiveness of different numbers of previous game events and results are shown in Table~\ref{tab:event-hp-testing}, where input transcript and chat time windows are set to 30 seconds, and the audio time window is set to 15 seconds. Increasing the number of previous game events improves the models' aggregate performance up until a specific threshold. However, it is observed that when this threshold is surpassed, there is a discernible decrement in performance. We hypothesise that the performance decline is due to the extended length of the previous events, which have less correlation with the target event.

\begin{table}[]
\centering
\resizebox{\columnwidth}{!}{%
\begin{tabular}{@{}c|c@{\hspace{8pt}}c@{\hspace{5pt}}c@{\hspace{5pt}}c@{\hspace{5pt}}|c@{\hspace{8pt}}c@{\hspace{5pt}}c@{\hspace{5pt}}c@{\hspace{5pt}}@{}}
\toprule
\multirow{2}{*}{\textbf{\begin{tabular}[c]{@{}c@{}}Game\\ Events\end{tabular}}} &
  \multicolumn{4}{c|}{\textbf{BERT}} &
  \multicolumn{4}{c}{\textbf{DeBERTaV3}} \\ \cline{2-9} 
 &
  \textbf{Kill} &
  \textbf{Tower} &
  \textbf{Dragon} &
  \textbf{All} &
  \textbf{Kill} &
  \textbf{Tower} &
  \textbf{Dragon} &
  \textbf{All} \\ \hline
3 & 85.39 & 29.73 & 18.84 & 63.32          & 88.64 & 25.23 & 53.62 & 69.26          \\
5 & 84.30 & 40.54 & 18.75 & \textbf{64.96} & 86.35 & 28.83 & 57.81 & 68.80          \\
7 & 80.58 & 40.91 & 21.67 & 62.95          & 85.25 & 42.73 & 56.67 & \textbf{70.98} \\
9 & 79.32 & 50.00 & 12.50 & 63.32          & 71.80 & 64.15 & 0.00  & 60.51          \\ \hline
\end{tabular}%
}

\caption{Hyperparameter testing on the Game Situation Understanding Model for different numbers of previous game events, where input transcript and chat time windows are 30s and the audio time window is 15s. A large number of previous game events may include less relevant histories and lead to a worse performance.}
\label{tab:event-hp-testing}
\end{table}

\begin{table}[]
\centering
\resizebox{\columnwidth}{!}{%
\begin{tabular}{@{}c|ccc|ccc@{}}
\toprule
\multirow{2}{*}{\textbf{Category}} & \multicolumn{3}{c|}{\textbf{GPT2}} & \multicolumn{3}{c}{\textbf{Pythia}} \\ \cmidrule(l){2-7} 
 & \textbf{Event} & \textbf{Coherence} & \textbf{Overall} & \textbf{Event} & \textbf{Coherence} & \textbf{Overall} \\ \midrule
\textbf{Kill} & 75.31\% & 75.31\% & 66.67\% & 24.69\% & 24.69\% & 33.33\% \\
\textbf{Tower} & 60.74\% & 59.26\% & 59.26\% & 39.26\% & 40.74\% & 40.74\% \\
\textbf{Dragon} & 61.62\% & 66.67\% & 59.60\% & 38.38\% & 33.33\% & 40.40\% \\ \midrule
\textbf{All} & 64.76\% & 65.71\% & 61.27\% & 35.24\% & 34.29\% & 38.73\% \\ \bottomrule
\end{tabular}%
}
\caption{Human evaluation comparison between GPT2 and Pythia commentaries. GPT2 gains better support from human annotators across all 3 aspects compared to Pythia.
}
\label{tab:human-eval}
\end{table}

\begin{figure*}[]
  \centering
  \includegraphics[width=1\textwidth]{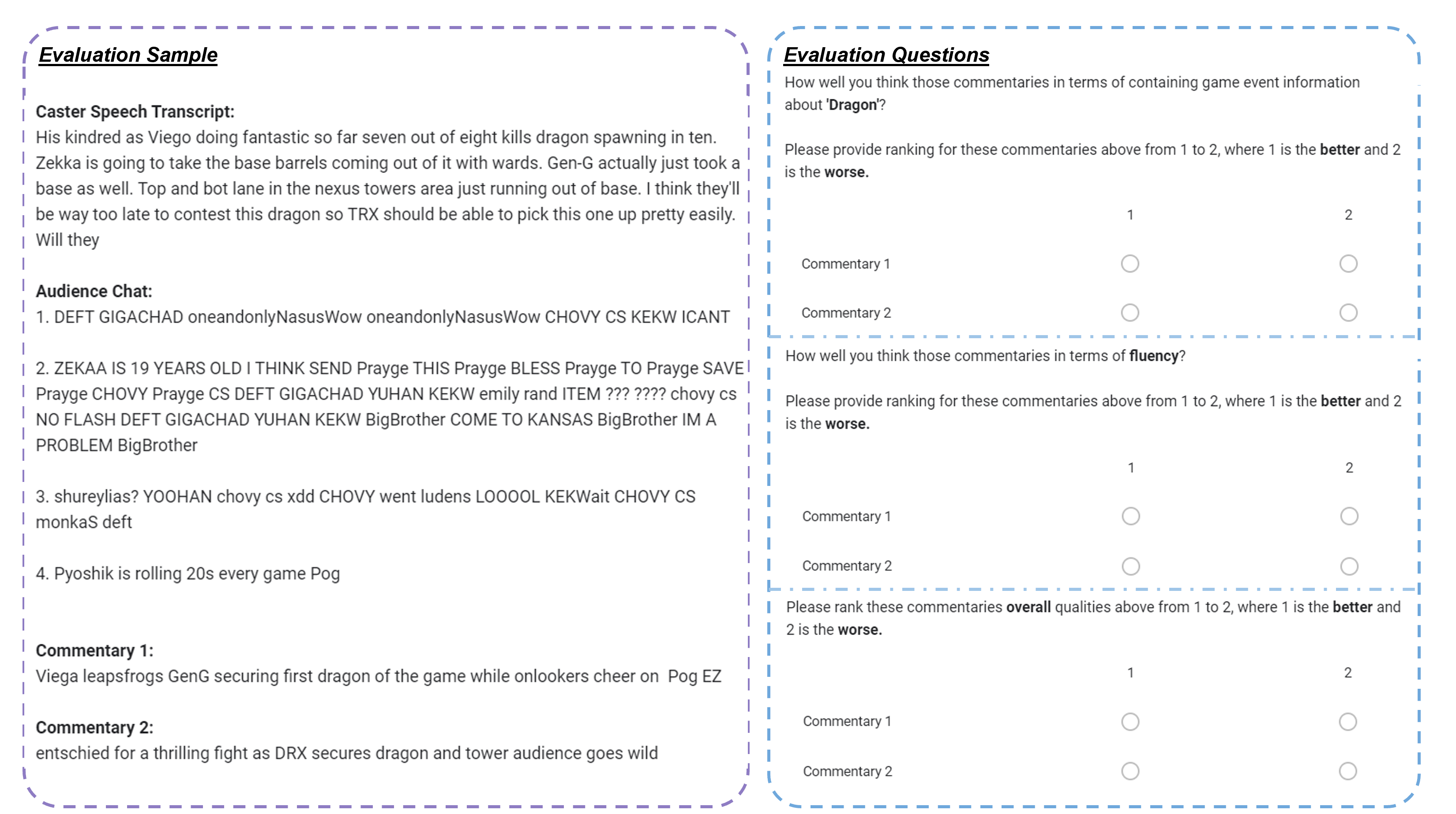}
  \caption{Screenshot of a human evaluation sample. Workers are shown the caster's speech with truncated audience chats on the top left. We provide the generated commentaries on the bottom left. The worker ranks these two commentaries in terms of the inclusion of the game event, coherence and overall quality. More evaluations samples can be found in Appendix B. }
  \label{fig:human-evaluation}
\end{figure*}

\noindent\textbf{Human Evaluation}
\label{sec:human-evaluation}
Automatic metrics may not correlate well with human judgments in different aspects~\cite{feqa}, therefore we conduct the human evaluation to enrich the comprehensiveness of the results. We recruited nine volunteers aged between 25 and 30, all holding at least a Bachelor's degree, to participate in the human evaluation. The group was composed of three females and six males, each with a general understanding of League of Legends. While one participant was a native English speaker, the other eight were proficient in English.  We randomly collected testing samples for evaluating the commentaries from GPT2 and Pythia by the nine workers, resulting in 1,890 instances of human feedback. 
For the human evaluation survey, participants were presented with the original transcript, the truncated chat, and the generated commentaries from the baseline models. They are then asked to rank the commentaries based on the following three criteria: 

\begin{itemize}
    \item \textbf{Game Event Information:} The quality of commentaries in terms of the game event-related expressions.
    \item \textbf{Coherence:} The quality of commentaries in terms of fluency and logic.
    \item \textbf{Overall}: The overall quality of commentaries regarding the above criteria and any other game-related criteria.
\end{itemize}

The sample evaluation questions are shown in Figure~\ref{fig:human-evaluation}.  As shown in Table~\ref{tab:human-eval}, commentaries of GPT2 are more preferred by humans in all categories which aligns with the results from automatic evaluation metrics.

\section{Conclusions}
We introduce GAME-MUG, a multimodal dataset tailored for understanding game situations and generating commentary in esports. By amalgamating diverse sources of game-related information, including game event logs, caster's speech transcripts, audience conversations, and game match audio, GAME-MUG offers a comprehensive repository that encapsulates the multifaceted nature of esports engagement.
Our proposed joint integration model represents a significant step forward in leveraging multimodal data for enhanced game understanding and commentary generation. By fusing multiple data modalities, our model demonstrates improved proficiency in comprehending intricate game dynamics, thereby facilitating the generation of more human-like and contextually rich commentary. By elucidating the interplay between game situations and emotional cues extracted from multimodal inputs, our model excels in capturing the essence of esports competition, fostering a deeper connection with the audience. Our decision to make GAME-MUG dataset publicly available will catalyse the development of practical applications and foster innovation within the esports community.

\clearpage
\bibliographystyle{ACM-Reference-Format}
\bibliography{sample-base}

\appendix

\section{Full Hyperparamemter Testing Results}
The complete hyperparameter results are displayed in Table~\ref{tab:hp-full}. We conducted experiments using 15s and 30s time windows for transcripts and chats, and 5s, 10s, and 15s time windows for audio. Additionally, we experimented with time-series events ranging from 3 to 10.
\begin{table*}[h]
\centering
\resizebox{\textwidth}{!}{%
\begin{tabular}{ccc|cccc|cccc|cccc|cccc}
\hline
\multirow{2}{*}{\textbf{\begin{tabular}[c]{@{}l@{}}Transcript\\ + Chat\end{tabular}}} &
  \multirow{2}{*}{\textbf{Audio}} &
  \multirow{2}{*}{\textbf{\begin{tabular}[c]{@{}l@{}}Game\\ Events\end{tabular}}} &
  \multicolumn{4}{c|}{\textbf{BERT}} &
  \multicolumn{4}{c|}{\textbf{RoBERTa}} &
  \multicolumn{4}{c|}{\textbf{DeBERTaV3}} &
  \multicolumn{4}{c}{\textbf{XLNet}} \\ \cline{4-19} 
 &
   &
   &
  \textbf{Kill} &
  \textbf{Tower} &
  \textbf{Dragon} &
  \textbf{All} &
  \textbf{Kill} &
  \textbf{Tower} &
  \textbf{Dragon} &
  \textbf{All} &
  \textbf{Kill} &
  \textbf{Tower} &
  \textbf{Dragon} &
  \textbf{All} &
  \textbf{Kill} &
  \textbf{Tower} &
  \textbf{Dragon} &
  \textbf{All} \\ \hline
15s & 5s  & 3  & 90.26 & 16.22 & 1.45  & 60.86 & 85.71 & 29.73 & 0.00  & 60.86 & 73.38 & 29.73 & 0.00  & 53.07 & 77.92 & 27.93 & 0.00  & 55.53 \\
15s & 5s  & 4  & 85.67 & 36.04 & 5.97  & 62.97 & 80.33 & 58.56 & 7.46  & 65.06 & 79.00 & 58.56 & 7.46  & 64.23 & 82.00 & 30.63 & 1.49  & 58.79 \\
15s & 5s  & 5  & 86.69 & 33.33 & 12.50 & 63.89 & 80.20 & 62.16 & 4.69  & 65.60 & 82.25 & 53.15 & 23.44 & 67.31 & 92.15 & 22.52 & 0.00  & 63.03 \\
15s & 5s  & 6  & 84.62 & 31.53 & 19.67 & 63.10 & 79.72 & 59.46 & 14.75 & 66.16 & 83.22 & 50.45 & 21.31 & 67.03 & 83.92 & 56.76 & 0.00  & 66.16 \\
15s & 5s  & 7  & 85.25 & 30.91 & 16.67 & 62.72 & 83.09 & 51.82 & 25.00 & 67.63 & 83.81 & 48.18 & 30.00 & 67.86 & 85.25 & 55.45 & 0.00  & 66.52 \\
15s & 5s  & 8  & 82.05 & 36.70 & 17.86 & 62.56 & 78.02 & 55.96 & 23.21 & 65.53 & 83.15 & 42.20 & 35.71 & 66.89 & 86.08 & 62.39 & 0.00  & 69.18 \\
15s & 5s  & 9  & 80.83 & 33.02 & 17.86 & 60.75 & 79.32 & 55.66 & 26.79 & 66.59 & 83.46 & 38.68 & 39.29 & 66.59 & 87.22 & 53.77 & 1.79  & 67.76 \\
15s & 5s  & 10 & 81.92 & 33.96 & 15.38 & 61.48 & 78.46 & 57.55 & 25.00 & 66.51 & 83.08 & 43.40 & 46.15 & 68.42 & 87.31 & 55.66 & 7.69  & 69.38 \\ \hline
15s & 10s & 3  & 86.04 & 30.63 & 4.35  & 61.89 & 69.16 & 57.66 & 7.25  & 57.79 & 71.43 & 61.26 & 4.35  & 59.63 & 65.58 & 33.33 & 13.04 & 50.82 \\
15s & 10s & 4  & 87.00 & 36.04 & 7.46  & 64.02 & 84.33 & 33.33 & 31.34 & 65.06 & 77.67 & 61.26 & 11.94 & 64.64 & 74.00 & 33.33 & 10.45 & 55.65 \\
15s & 10s & 5  & 84.98 & 37.84 & 12.50 & 63.89 & 79.52 & 55.86 & 28.12 & 66.88 & 81.91 & 49.55 & 17.19 & 65.38 & 83.96 & 42.34 & 7.81  & 63.68 \\
15s & 10s & 6  & 81.82 & 34.23 & 21.31 & 62.23 & 79.72 & 56.76 & 24.59 & 66.81 & 81.12 & 44.14 & 40.98 & 66.81 & 79.37 & 54.05 & 1.64  & 62.88 \\
15s & 10s & 7  & 83.09 & 30.91 & 13.33 & 60.94 & 80.94 & 46.36 & 35.00 & 66.29 & 84.17 & 45.45 & 35.00 & 68.08 & 84.53 & 55.45 & 3.33  & 66.52 \\
15s & 10s & 8  & 81.68 & 36.70 & 16.07 & 62.10 & 78.75 & 52.29 & 28.57 & 65.75 & 84.98 & 43.12 & 39.29 & 68.72 & 86.45 & 58.72 & 10.71 & 69.86 \\
15s & 10s & 9  & 82.33 & 31.13 & 19.64 & 61.45 & 79.70 & 57.55 & 23.21 & 66.82 & 81.95 & 40.57 & 48.21 & 67.29 & 86.47 & 50.00 & 7.14  & 67.06 \\
15s & 10s & 10 & 80.38 & 35.85 & 15.38 & 61.00 & 82.31 & 58.49 & 17.31 & 68.18 & 82.69 & 42.45 & 44.23 & 67.70 & 86.54 & 57.55 & 3.85  & 68.90 \\ \hline
15s & 15s & 3  & 82.14 & 36.94 & 2.90  & 60.66 & 70.13 & 65.77 & 2.90  & 59.63 & 80.84 & 54.05 & 11.59 & 64.96 & 59.42 & 60.36 & 8.70  & 52.46 \\
15s & 15s & 4  & 84.33 & 44.14 & 8.96  & 64.44 & 75.67 & 63.06 & 5.97  & 62.97 & 80.00 & 54.05 & 23.88 & 66.11 & 63.00 & 53.15 & 14.93 & 53.97 \\
15s & 15s & 5  & 83.28 & 39.64 & 10.94 & 63.03 & 74.74 & 69.37 & 6.25  & 64.10 & 85.32 & 36.94 & 34.38 & 66.88 & 73.38 & 66.67 & 4.69  & 62.39 \\
15s & 15s & 6  & 81.82 & 38.74 & 13.11 & 62.23 & 81.12 & 54.95 & 26.23 & 67.47 & 84.27 & 49.55 & 34.43 & 69.21 & 76.92 & 66.67 & 3.28  & 64.63 \\
15s & 15s & 7  & 83.45 & 30.91 & 15.00 & 61.38 & 80.22 & 50.91 & 28.33 & 66.07 & 86.33 & 37.27 & 36.67 & 67.63 & 82.01 & 56.36 & 1.67  & 64.96 \\
15s & 15s & 8  & 79.12 & 38.53 & 14.29 & 60.73 & 77.29 & 61.47 & 26.79 & 66.89 & 87.18 & 43.12 & 41.07 & 70.32 & 83.15 & 62.39 & 3.57  & 67.81 \\
15s & 15s & 9  & 81.58 & 33.02 & 8.93  & 60.05 & 77.44 & 58.49 & 19.64 & 65.19 & 81.95 & 41.51 & 42.86 & 66.82 & 85.71 & 52.83 & 3.57  & 66.82 \\
15s & 15s & 10 & 80.00 & 37.74 & 17.31 & 61.48 & 81.92 & 56.60 & 21.15 & 67.94 & 86.54 & 38.68 & 44.23 & 69.14 & 84.62 & 53.77 & 11.54 & 67.70 \\ \hline
30s & 5s  & 3  & 80.52 & 34.23 & 20.29 & 61.48 & 81.49 & 48.65 & 44.93 & 68.85 & 82.47 & 32.43 & 71.01 & 69.47 & 85.71 & 54.05 & 10.14 & 67.83 \\
30s & 5s  & 4  & 81.33 & 43.24 & 23.88 & 64.44 & 79.00 & 51.35 & 40.30 & 67.15 & 86.00 & 44.14 & 40.30 & 69.87 & 84.33 & 51.35 & 17.91 & 67.36 \\
30s & 5s  & 5  & 80.55 & 42.34 & 25.00 & 63.89 & 81.91 & 50.45 & 37.50 & 68.38 & 83.62 & 35.14 & 57.81 & 68.59 & 82.94 & 50.45 & 23.44 & 67.09 \\
30s & 5s  & 6  & 83.22 & 43.24 & 26.23 & 65.94 & 80.42 & 55.86 & 36.07 & 68.56 & 83.57 & 31.53 & 57.38 & 67.47 & 84.27 & 47.75 & 21.31 & 67.03 \\
30s & 5s  & 7  & 81.29 & 41.82 & 25.00 & 64.06 & 80.58 & 57.27 & 30.00 & 68.08 & 83.81 & 40.00 & 56.67 & 69.42 & 84.53 & 48.18 & 25.00 & 67.63 \\
30s & 5s  & 8  & 79.12 & 46.79 & 14.29 & 62.79 & 81.68 & 56.88 & 33.93 & 69.41 & 83.15 & 41.28 & 57.14 & 69.41 & 84.98 & 50.46 & 17.86 & 67.81 \\
30s & 5s  & 9  & 78.57 & 46.23 & 14.29 & 62.15 & 77.82 & 51.89 & 32.14 & 65.42 & 68.42 & 36.79 & 0.00  & 51.64 & 86.47 & 46.23 & 19.64 & 67.76 \\
30s & 5s  & 10 & 78.85 & 52.83 & 9.62  & 63.64 & 81.15 & 55.66 & 30.77 & 68.42 & 71.15 & 71.70 & 0.00  & 62.44 & 85.00 & 46.23 & 21.15 & 67.22 \\ \hline
30s & 10s & 3  & 81.17 & 35.14 & 23.19 & 62.50 & 81.49 & 48.65 & 47.83 & 69.26 & 83.77 & 35.14 & 56.52 & 68.85 & 83.77 & 55.86 & 15.94 & 67.83 \\
30s & 10s & 4  & 82.33 & 40.54 & 25.37 & 64.64 & 77.33 & 49.55 & 43.28 & 66.11 & 85.67 & 30.63 & 50.75 & 67.99 & 82.67 & 52.25 & 17.91 & 66.53 \\
30s & 10s & 5  & 83.62 & 37.84 & 23.44 & 64.53 & 81.23 & 50.45 & 35.94 & 67.74 & 83.28 & 35.14 & 59.38 & 68.59 & 83.96 & 53.15 & 17.19 & 67.52 \\
30s & 10s & 6  & 81.82 & 45.05 & 22.95 & 65.07 & 79.02 & 54.95 & 32.79 & 67.03 & 83.57 & 38.74 & 59.02 & 69.43 & 85.31 & 52.25 & 16.39 & 68.12 \\
30s & 10s & 7  & 80.94 & 39.09 & 25.00 & 63.17 & 82.73 & 48.18 & 31.67 & 67.41 & 83.09 & 39.09 & 56.67 & 68.75 & 83.09 & 49.09 & 20.00 & 66.29 \\
30s & 10s & 8  & 79.12 & 51.38 & 14.29 & 63.93 & 81.32 & 53.21 & 35.71 & 68.49 & 83.88 & 42.20 & 55.36 & 69.86 & 84.62 & 51.38 & 19.64 & 68.04 \\
30s & 10s & 9  & 78.20 & 43.40 & 12.50 & 60.98 & 80.83 & 50.94 & 32.14 & 67.06 & 69.17 & 74.53 & 0.00  & 61.45 & 87.97 & 45.28 & 21.43 & 68.69 \\
30s & 10s & 10 & 80.00 & 49.06 & 11.54 & 63.64 & 80.77 & 56.60 & 30.77 & 68.42 & 74.62 & 69.81 & 0.00  & 64.11 & 86.92 & 48.11 & 23.08 & 69.14 \\ \hline
30s & 15s & 3  & 85.39 & 29.73 & 18.84 & 63.32 & 81.17 & 48.65 & 36.23 & 67.42 & 88.64 & 25.23 & 53.62 & 69.26 & 79.55 & 54.95 & 24.64 & 66.19 \\
30s & 15s & 4  & 83.67 & 37.84 & 23.88 & 64.64 & 79.67 & 46.85 & 32.84 & 65.48 & 86.67 & 33.33 & 52.24 & 69.46 & 83.00 & 47.75 & 20.90 & 66.11 \\
30s & 15s & 5  & 84.30 & 40.54 & 18.75 & 64.96 & 82.59 & 49.55 & 32.81 & 67.95 & 86.35 & 28.83 & 57.81 & 68.80 & 82.94 & 53.15 & 23.44 & 67.74 \\
30s & 15s & 6  & 83.57 & 43.24 & 18.03 & 65.07 & 80.42 & 52.25 & 31.15 & 67.03 & 86.71 & 31.53 & 59.02 & 69.65 & 83.92 & 53.15 & 18.03 & 67.69 \\
30s & 15s & 7  & 80.58 & 40.91 & 21.67 & 62.95 & 83.81 & 52.73 & 31.67 & 69.20 & 85.25 & 42.73 & 56.67 & 70.98 & 82.73 & 50.91 & 20.00 & 66.52 \\
30s & 15s & 8  & 79.85 & 49.54 & 10.71 & 63.47 & 79.12 & 55.05 & 30.36 & 66.89 & 86.81 & 39.45 & 53.57 & 70.78 & 83.15 & 51.38 & 25.00 & 67.81 \\
30s & 15s & 9  & 79.32 & 50.00 & 12.50 & 63.32 & 80.83 & 51.89 & 35.71 & 67.76 & 71.80 & 64.15 & 0.00  & 60.51 & 86.47 & 48.11 & 21.43 & 68.46 \\
30s & 15s & 10 & 79.62 & 51.89 & 11.54 & 64.11 & 80.00 & 56.60 & 32.69 & 68.18 & 69.62 & 69.81 & 5.77  & 61.72 & 83.85 & 50.94 & 25.00 & 68.18 \\ \hline
\end{tabular}%
}
\caption{Full hyperparameter testing results.}
\label{tab:hp-full}
\end{table*}

\section{Qualitative Samples}
In this Section we show more qualitative samples of the generated commentary by our Game-MUG joint learning framework.

\begin{figure*}[t!]
    \centering
    \begin{subfigure}{}
        \centering
        \includegraphics[width=1\textwidth]{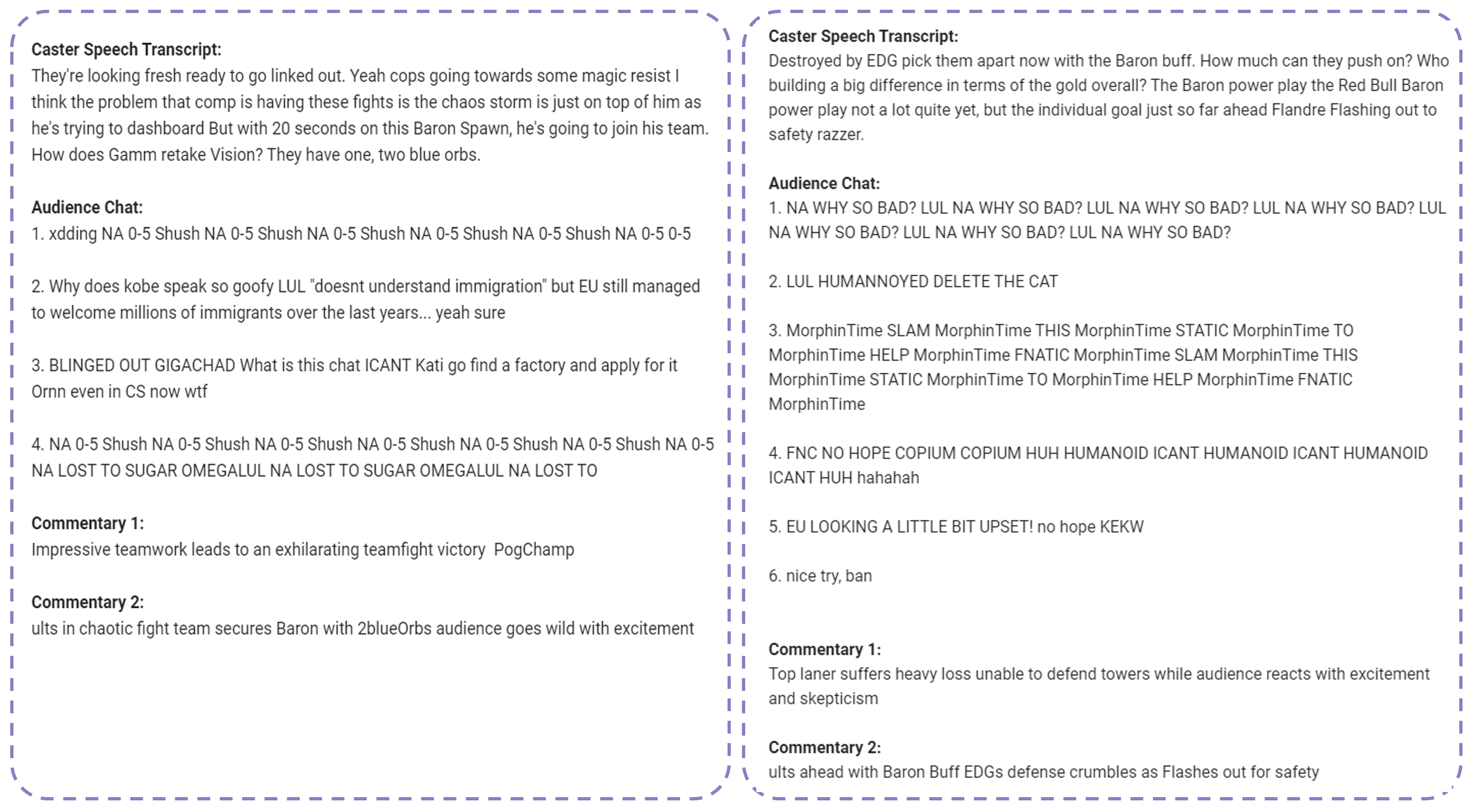}
    \end{subfigure}%
   \vfill
    \begin{subfigure}{}
        \centering
        \includegraphics[width=1\textwidth]{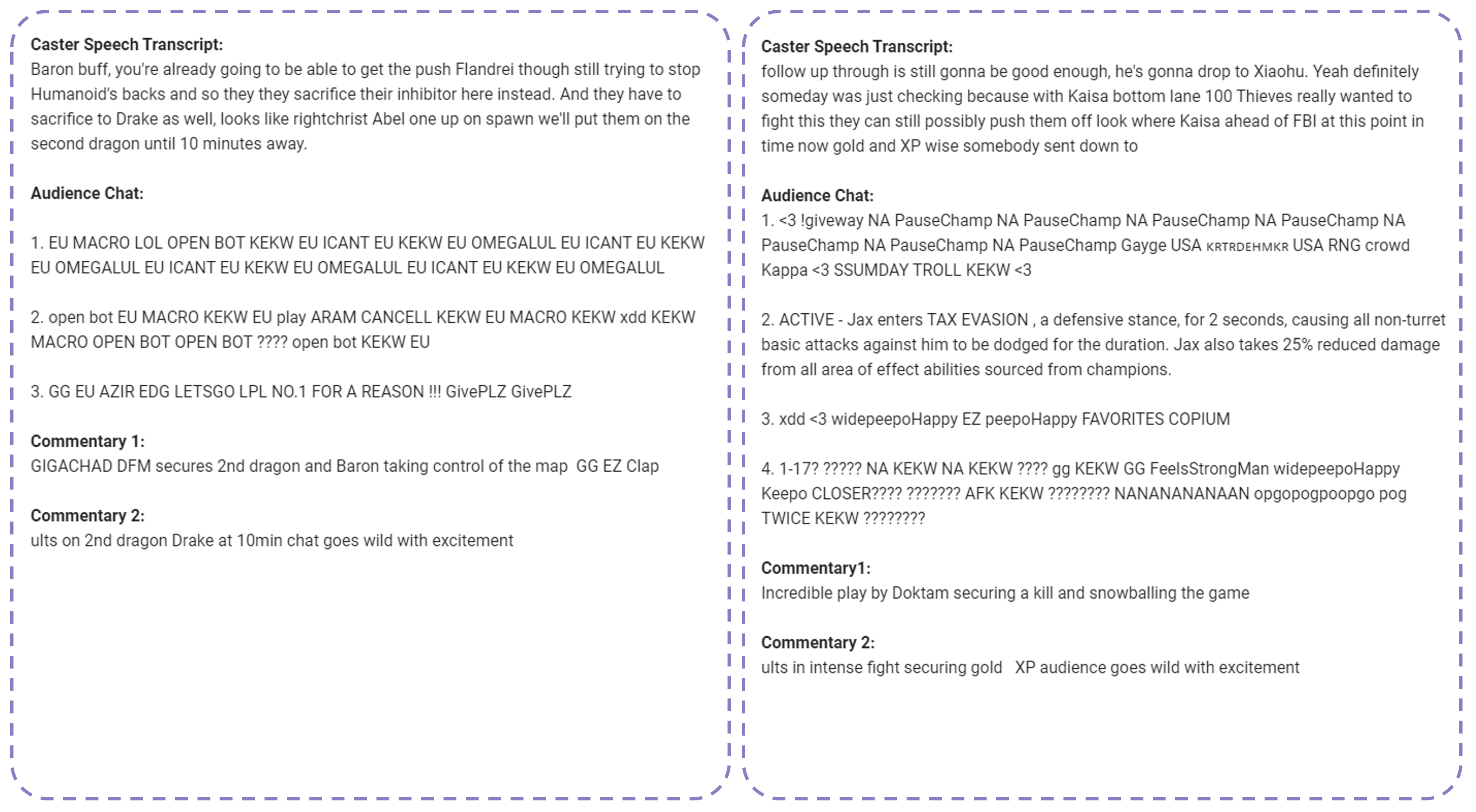}
    \end{subfigure}
    \caption{Screenshot of the commentary samples. }
  \label{fig:gamemug_appendix_sample12}
\end{figure*}

\begin{figure*}[t!]
    \centering
    \begin{subfigure}{}
        \centering
        \includegraphics[width=1\textwidth]{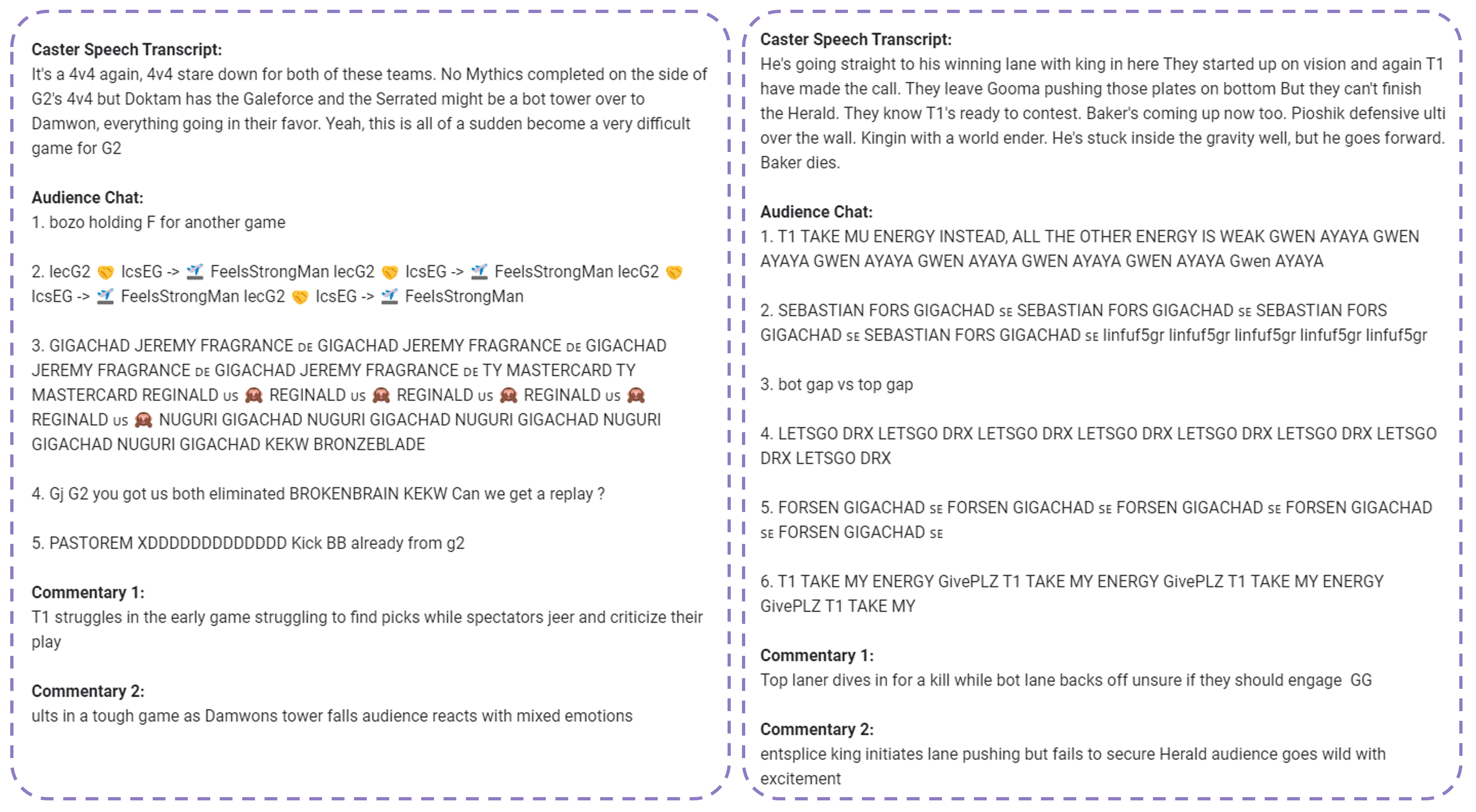}
    \end{subfigure}%
    \vfill
    \begin{subfigure}{}
        \centering
        \includegraphics[width=1\textwidth]{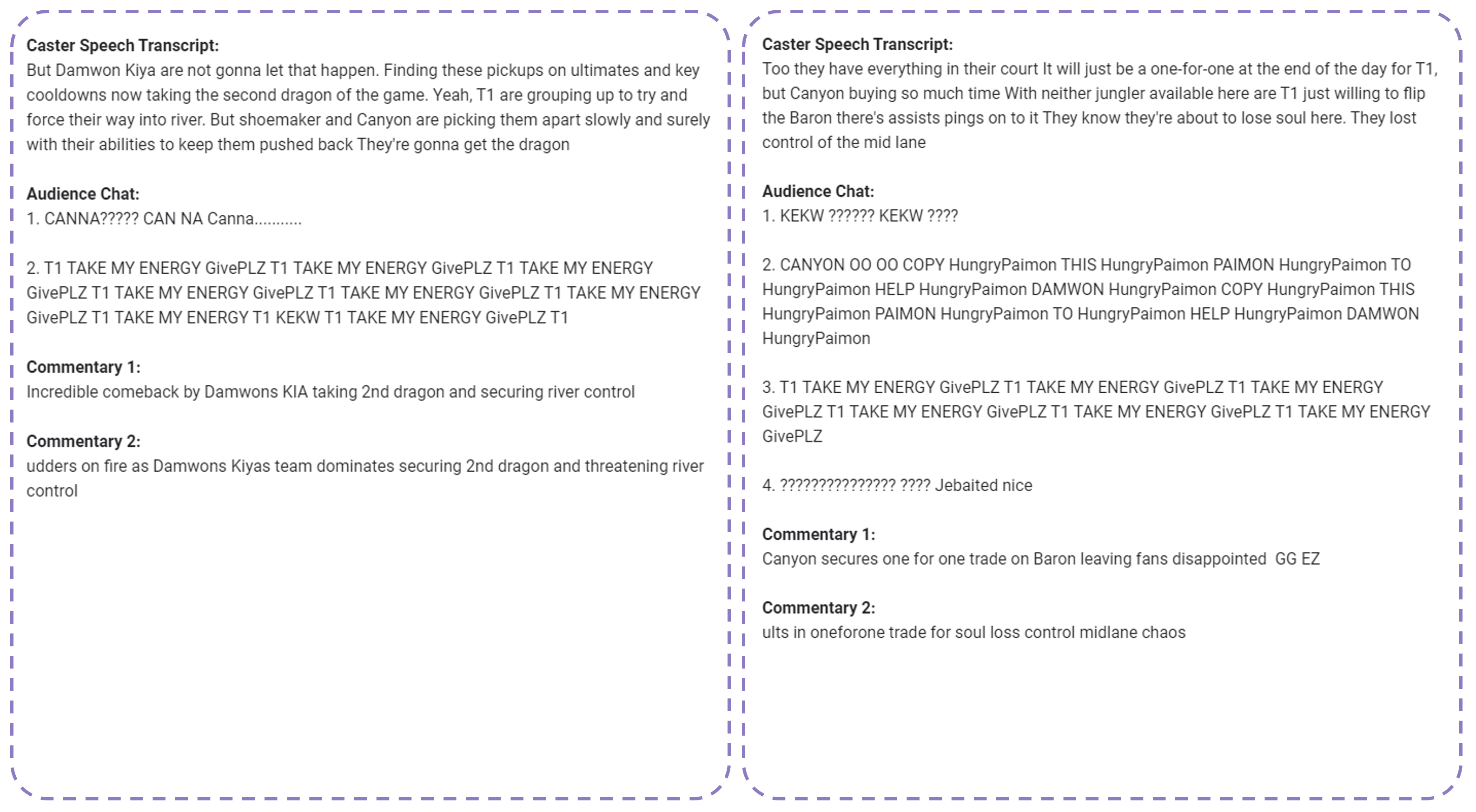}
    \end{subfigure}
    \caption{Screenshot of the commentary samples (continued). }
  \label{fig:gamemug_appendix_sample34}
\end{figure*}

\begin{figure*}[]
  \centering
  \includegraphics[width=1\textwidth]{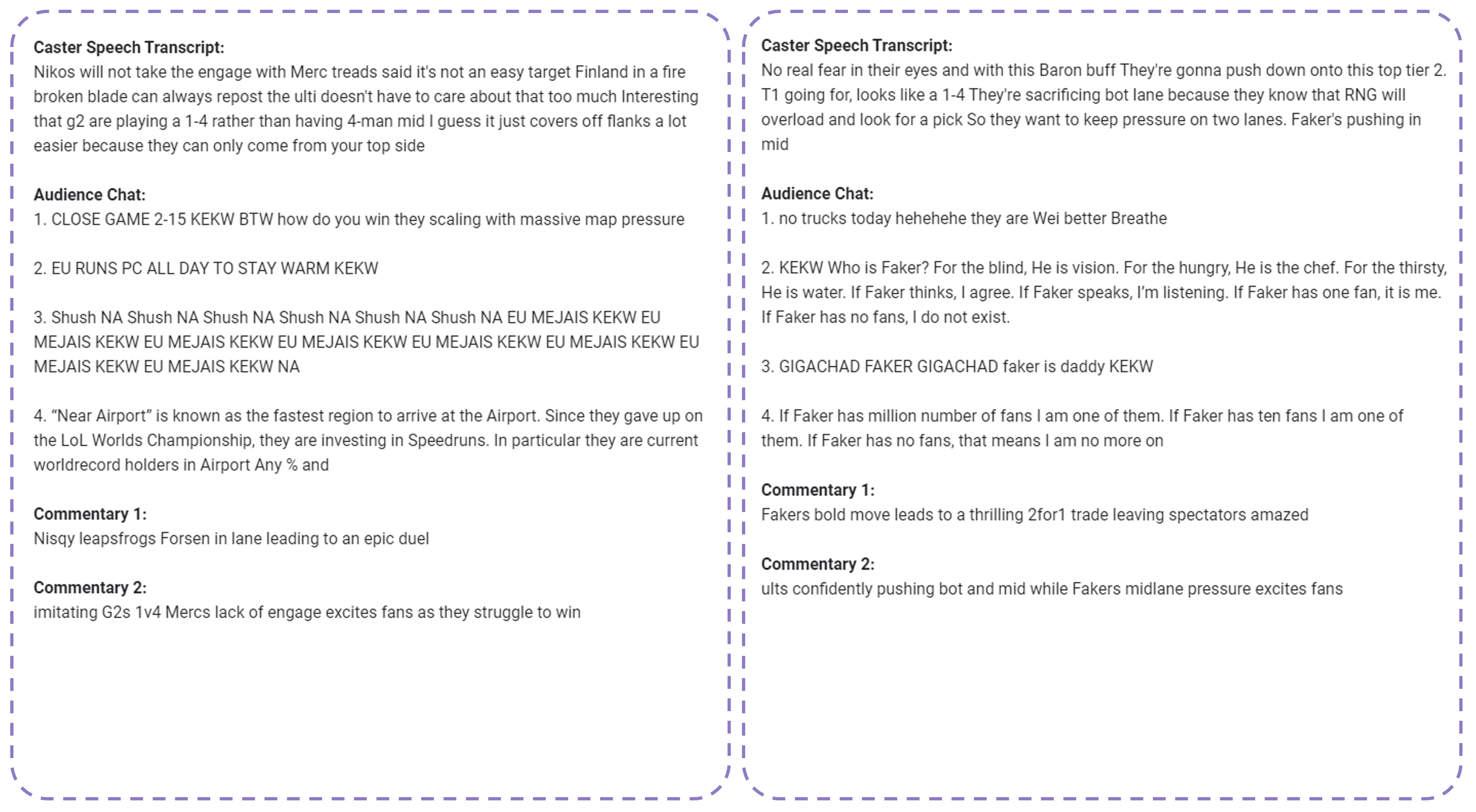}
  \caption{Screenshot of the commentary samples (continued).}
  \label{fig:gamemug_appendix_sample5}
\end{figure*}

\end{document}